\documentclass[a4paper]{article}
\usepackage{INTERSPEECH2019,amsmath,graphicx, subcaption}
\usepackage{multirow}
\usepackage{hhline}
\usepackage{color}
\usepackage{flushend}
\usepackage{booktabs}
\usepackage{enumitem}
\usepackage{amsmath}

\newcommand{\qiantong}[1]{{\color{red}  }}

\DeclareMathOperator*{\logadd}{logadd}
\title{Fully Convolutional Speech Recognition}
\name{\begin{tabular}{c}Neil Zeghidour$^{1,2, *}$, Qiantong Xu$^{1, *}$, Vitaliy Liptchinsky$^1$, Nicolas Usunier$^1$, \\ Gabriel Synnaeve$^1$, Ronan Collobert$^1$ \end{tabular}\thanks{* Equal contribution}}
\address{$^1$ Facebook A.I. Research, Paris, France; New York \& Menlo Park, USA\\
$^2$ CoML, ENS/CNRS/EHESS/INRIA/PSL Research University, Paris, France}
\email{}

\begin{document}
%
\maketitle
\begin{abstract}

Current state-of-the-art speech recognition systems build on recurrent
neural networks for acoustic and/or language modeling, and rely on
feature extraction pipelines to extract mel-filterbanks or cepstral
coefficients. In this paper we present an alternative approach based
solely on convolutional neural networks, leveraging recent advances in
acoustic models from the raw waveform and language modeling. This
fully convolutional approach is trained end-to-end to predict
characters from the raw waveform, removing the feature extraction step
altogether. An external convolutional language model is used to decode words. On Wall Street Journal, our model matches the current state-of-the-art. On Librispeech, we report state-of-the-art performance among end-to-end models, including Deep Speech 2, that was trained with 12 times more acoustic data and significantly more linguistic data.

\end{abstract}

\section{Introduction}
\label{sec:intro}
Recent work on convolutional neural network architectures shows they are competitive with recurrent architectures even on tasks where modeling long-range dependencies is critical, such as language modeling \cite{glu}, machine translation \cite{fairseq,fairseq_translation} and speech synthesis \cite{wavenet}. In end-to-end speech recognition however, recurrent neural networks are still prevalent for acoustic and/or language modeling \cite{CTC,RNNLM,deepspeech2,seq2seqspeech, improvedzeyer}.

There is a history of using convolutional networks in speech recognition, but only as part of an otherwise more traditional pipeline. They were first introduced as TDNNs to predict phoneme classes \cite{tdnn}, and later to generate HMM posteriorgrams \cite{cnn_abdelhamid}. They have recently been used in end-to-end systems, but only in combination with recurrent layers \cite{deepspeech2}, or n-gram language models  \cite{wav2letter2}, or for phone recognition \cite{zhang_towards, tdfbanks1}. Convolutional architectures are prevalent when learning from the raw waveform  \cite{palaz2015convolutional,hoshen2015speech,sainath2015learning,tdfbanks1,tdfbanks2}, because they naturally model the computation of standard features such as mel-filterbanks. Given the evidence that convolutional networks are also suitable on long-range dependency tasks,
we expect them to be competitive at all levels of the speech recognition pipeline.

In this paper, we present a fully convolutional approach to end-to-end speech recognition. Building on recent advances in convolutional learnable front-ends for speech \cite{tdfbanks1,tdfbanks2}, convolutional acoustic models \cite{wav2letter2}, and convolutional language models \cite{glu}, our model is a deep convolutional network that takes the raw waveform as input and is trained end-to-end to predict letters. Sentences are then predicted using beam-search decoding with a convolutional language model. 

In addition to presenting the first application of convolutional language models to speech recognition, the main contribution of the paper is to show that fully convolutional architectures achieve state-of-the-art performance among end-to-end systems. Thus, our results challenge the prevalence of recurrent architectures for speech recognition, and they parallel the prior results on other application domains that convolutional architectures are on-par with recurrent ones.

More precisely, we perform experiments on the large vocabulary task of the Wall Street Journal dataset (WSJ) and on the 1000h Librispeech. Our overall pipeline improves the state-of-the-art of end-to-end systems on both datasets. In particular, we decrease by 2\%  (absolute) the Word Error Rate on the noisy test set of Librispeech compared to DeepSpeech 2 \cite{deepspeech2} and the best sequence-to-sequence model \cite{improvedzeyer}. On clean speech, the improvement is about ~0.5\% 
on Librispeech compared to the best end-to-end systems; on WSJ, our results are competitive with the current state-of-the-art, a DNN-HMM system \cite{chan2015deep}. 

In particular, the detailed results show that the convolutional language model yields systematic and consistent improvement over a $4$-gram language model for its better perplexity and larger receptive field. In addition, we complement the promising results of \cite{tdfbanks2} regarding the performance of learning the front-end of speech recognition systems: first, we show that learning the front-end yields substantial improvements on noisy speech compared to a mel-filterbanks front-end. Second, we show additional improvements on both WSJ and Librispeech by varying the number of filters in the learnable front-end, leading to a 1.5\% absolute decrease in WER on the noisy test set of Librispeech. Our results are the first in which an end-to-end system trained on the raw waveform achieves state-of-the-art performance (among all end-to-end systems) on two datasets. 

\begin{figure*}
\includegraphics[width=\textwidth]{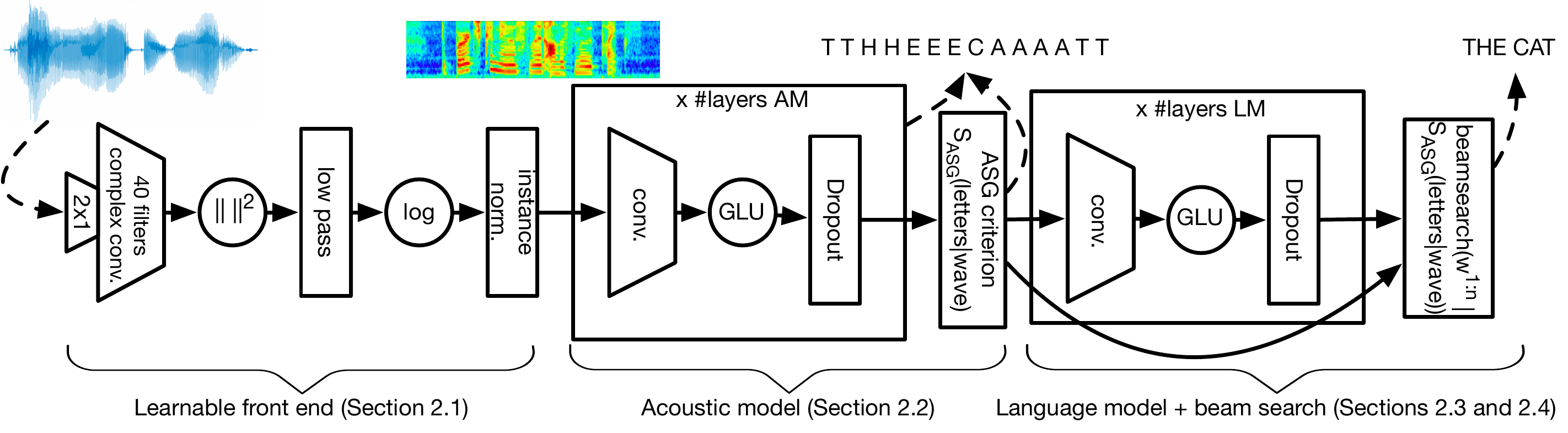}
\caption{Overview of the fully convolutional architecture.}
\label{fig:arch}
\end{figure*}
\section{Model}

Our approach, described in this section, is illustrated in Fig.~\ref{fig:arch}.

\label{sec:model}
\subsection{Convolutional Front-end}

Several proposals to learn the front-end of speech recognition systems have been made \cite{hoshen2015speech,sainath2015learning,tdfbanks1,tdfbanks2}. Following the comparison in \cite{tdfbanks2}, we consider their best architecture, called "scattering based" (hereafter refered to as \emph{learnable front-end}). The learnable front-end contains first a convolution of width 2 that emulates the pre-emphasis step used in mel-filterbanks. It is followed by a complex convolution of width $25$ms and $k$ filters. After taking the squared absolute value, a low-pass filter of width $25$ms and stride $10$ms performs de\-ci\-ma\-tion. The front-end finally applies a log-compression and a per-channel mean-variance normalization (equivalent to an instance normalization layer \cite{ulyanov2017instance}). Following \cite{tdfbanks2}, the "pre-emphasis" convolution is initialized to $[-0.97;1]$, and then trained with the rest of the network. The low-pass filter is kept constant to a squared Hanning window, and the complex convolutional layer is initialized randomly. 

In addition to the $k=40$ filters used by \cite{tdfbanks2}, we experiment with $k=80$ filters. Notice that since the stride is the same as for mel-filterbanks, acoustic models on top of the learnable front-ends can also be applied to mel-filterbanks (simply modifying the number of input channels if $k\neq 40$).

\subsection{Convolutional Acoustic Model}
The acoustic model is a convolutional neural network with gated linear units \cite{glu}, which is fed with the output of the learnable front-end. As in \cite{wav2letter2}, the networks uses a growing number of channels, and dropout \cite{dropout} for regula\-rization. These acoustic models are trained to predict letters directly with the Auto Segmentation Criterion (ASG) \cite{wav2letter}. 
The ASG criterion is similar to CTC \cite{CTC} except that it adds input-independent transition scores between letters.
The depth, number of feature maps per layer, receptive field and amount of dropout of models on each dataset are adjusted individually based on the amount of training data.

\subsection{Convolutional Language Model}
\label{sec:lm}
The convolutional language model (LM) is the GCNN-14B from \cite{glu}, which achieved competitive results on language modeling benchmarks with similar size of vocabulary and training data to the ones used in our experiments. The network contains 14 convolutional residual blocks \cite{resnet}, and this deep architecture gives us a large enough receptive field. In each residual block, two $1 \times 1$ 1-D convolutional layers are placed at the beginning and the end serving as bottlenecks for computational efficiency. Gated linear units are used as activation functions. 

We use the language model to score candidate transcriptions in addition to the acoustic model in the beam search decoder described in the next section. Compared to n-gram LMs, convolutional LMs allow the decoder to look at longer context with better perplexity.

\subsection{Beam-search decoder}
\label{sec:beam}
We use the beam-search decoder presented in \cite{wav2letter2} to generate word sequences given the output from our acoustic model. Given input $X$ to the acoustic model, the decoder finds the word transcription $W$ which maximizes:
\begin{equation}
\label{eq:beamscore}
\mathit{AM} (W|X) + \alpha ~log~ P_{lm}(W) + \beta | W | - \gamma | \{ i | \pi_i = \left<sil\right> \}|,
\end{equation}
where $\pi$ is a path representing a valid sequence of letters for $W$ and $\pi_i$ is the $i$-th letter in this sequence.
The score of the acoustic model is computed based on the score of paths of letters (including silences) that are compatible with the output sequence. Denoting by $\mathcal{G}_{\rm asg}$ the corresponding graph, the score of a sequence of words $W$ given by the accoustic model is

\begin{equation}
\mathit{AM} (W) = \logadd_{\pi \in \mathcal{G}_{\rm asg}(W)} \sum_{t = 1}^T f_{\pi_t}^t + g_{\pi_{t-1}, \pi_t},
\end{equation}
where $T$ is the number of output frames to be decoded, $f_i^t$ is the log-probability of letter $i$ for frame $t$ and and $g_{i,j}$ is the transition score from letter $i$ to letter $j$, respectively.

The other hyper-parameters $\alpha,\beta,\gamma\geq 0$ in \eqref{eq:beamscore} control the weight of the language model, the word insertion reward, and the silence insertion penalty. Additional hyper-parameters of the beam search are the beam size and the beam score. The latter is a global threshold on the score of a hypothesis to be included in the beam and is only used for efficiency.

To make decoding more efficient, at each time step, we merge all hypotheses getting into the same node in $\mathcal{G}_{asg}$ before sorting so as to better utilize the beam. Also, to reduce the number of forward passes through the LM, we cache the scores of unchanged LM states from previous time steps, and only forward batches of new hypotheses generated at current step.

\section{Experiments}
\label{sec:experiments}
We evaluate our approach on the large vocabulary task of the Wall Street Journal (WSJ) dataset \cite{wsj}, which contains 80 hours of clean read speech, and Librispeech \cite{librispeech}, which contains $1000$ hours with separate train/dev/test splits for clean and noisy speech. Each dataset comes with official textual data to train language models, which contain $37$ million tokens for WSJ, $800$ million tokens for Librispeech. Our language models are trained separately for each dataset on the official text data only.
These datasets were chosen to study the impact of the different components of our system at different scales of training data and in different recording conditions. 

The models are evaluated in Word Error Rate (WER). Our experiments use the open source codes of wav2letter\footnote{https://github.com/facebookresearch/wav2letter} for the acoustic model, and fairseq\footnote{https://github.com/facebookresearch/fairseq} for the language model. More details on the experimental setup are given below. 

{\bf Baseline} Our baseline for each dataset follows \cite{wav2letter2}. It uses the same convolutional acoustic model as our approach but a mel-filterbanks front-end and a $4$-gram language model.

{\bf Training/test splits~} On WSJ, models are trained on \textit{si284}. \textit{nov93dev} is used for validation and \textit{nov92} for test. On Librispeech, we train on the concatenation of \textit{train-clean} and {\it train-other}. The validation set is {\it dev-clean} when testing on {\it test-clean}, and {\it dev-other} when testing on {\it test-other}.

{\bf Acoustic modeling} The architecture for the convolutional acoustic model is the "high dropout" model from \cite{wav2letter2} for Librispeech, which has 19 layers in addition to the front-end (mel-filterbanks for the baseline, or the learnable front-end for our approach). On WSJ, we use the lighter version used in \cite{tdfbanks2}, which has 17 layers. Dropout is applied at each layer after the front-end, following \cite{wav2letter}. The learnable front-end uses $40$ or $80$ filters. 

The acoustic models are trained following \cite{wav2letter2,tdfbanks2}, using SGD with a decreasing learning rate, weight normalization and gradient clipping at 0.2 and a momentum of 0.9.

{\bf Language modeling} As described in Section~\ref{sec:lm}, we use the GCNN-14B model of \cite{glu} with dropout at each convolutional and linear layer on both WSJ and Librispeech. The kernel size of the 1-D convolutional layer in the middle of each residual block is set to 5. In terms of the training vocabulary, we keep all the words (162K) in the WSJ training corpus, while only the most frequent 200K tokens out of 900K are kept for Librispeech. 
Language models are trained with Nesterov accelerated gradient \cite{nag}. Following \cite{glu}, we also use weight normalization and gradient clipping as regularization.

{\bf Hyperparameter tuning}
The parameters of the beam search (see Section \ref{sec:beam}) $\alpha$, $\beta$ and $\gamma$ are tuned on the validation set with a beam size of $2500$ and a beam score of $26$ for computational efficiency. Once $\alpha,\beta,\gamma$ are chosen, the test WER together with the optimal number on the dev set are further pushed with a beam size of $3000$ and a beam score of $50$.

\begingroup
\setlength{\tabcolsep}{3pt}
\begin{table}[]
    \centering
\begin{tabular}{lcccc} 
\toprule
\multicolumn{3}{c}{Model}&dev93&nov92\\
\midrule
\multicolumn{3}{l}{E2E Lattice-free MMI \cite{latticefreemmi}} & - & 4.1\\
\multicolumn{3}{l}{\small \emph{(data augmentation)}}& & \\
\multicolumn{3}{l}{CNN-DNN-BLSTM-HMM \cite{chan2015deep}}&6.6&3.5\\
\multicolumn{3}{l}{\small \emph{(speaker adaptation, 3k acoustic states)}}& &\\
\multicolumn{3}{l}{DeepSpeech 2 \cite{deepspeech2}}&5&3.6\\
\multicolumn{3}{l}{\small \emph{(12k training hours AM, common crawl LM)}}& &\\
\midrule
\multicolumn{2}{c}{Front-end}&{\centering{LM}}&&\\
\cmidrule{1-3}
\multicolumn{2}{l}{Mel-filterbanks} & 4-gram & 9.5 & 5.6 \\
\multicolumn{2}{l}{Mel-filterbanks} & ConvLM &7.5 &4.1\\
\multicolumn{2}{l}{Learnable front-end (40 filters)} & ConvLM &6.9 &3.7\\
\multicolumn{2}{l}{Learnable front-end (80 filters)} & ConvLM &6.8&3.5\\
\bottomrule
\end{tabular}
   \caption{WER (\%) on the open vocabulary task of WSJ.}
    \label{tab:wsj}
\vspace{-4mm}
\end{table}
\endgroup

\section{Results}
\label{sec:results}

\subsection{Word Error Rate results}

\subsubsection{Wall Street Journal dataset}
Table \ref{tab:wsj} shows Word Error Rates (WER) on WSJ for the current state-of-the-art and our models. The current best model trained on this dataset is an HMM-based system which uses a combination of convolutional, recurrent and fully connected layers, as well as speaker adaptation, and reaches $3.5\%$ WER on \textit{nov92}. DeepSpeech 2 shows a WER of $3.6\%$ but uses 150 times more training data for the acoustic model and huge text datasets for LM training. Finally, the state-of-the-art among end-to-end systems trained only on WSJ, and hence the most comparable to our system, uses lattice-free MMI on augmented data (with speed perturbation) and gets $4.1\%$ WER. Our baseline system, trained on mel-filterbanks, and decoded with a n-gram language model has a $5.6\%$ WER. Replacing the n-gram LM by a convolutional one reduces the WER to $4.1\%$, and puts our model on par with the current best end-to-end system. Replacing the speech features by a learnable front-end finally reduces the WER to $3.7\%$ and then to $3.5\%$ when doubling the number of learnable filters, improving over DeepSpeech 2 and matching the performance of the best HMM-DNN system.

\begin{table}[]
    \centering
    \hspace*{-0.5cm}
\begin{tabular}{lcccccc}
\toprule
\multicolumn{3}{c}{\multirow{2}{*}{Model}}                                      & \multicolumn{2}{c}{Dev WER} & \multicolumn{2}{c}{Test WER} \\ \cmidrule{4-7} 
\multicolumn{3}{c}{}                                                            & Clean        & Other        & Clean         & Other        \\ \midrule
\multicolumn{3}{l}{CAPIO (single) \cite{capio}}   &  3.02 & 8.28 & 3.56 & 8.58              \\
\multicolumn{3}{l}{\small \emph{(DNN-HMM, speaker adapt.)}}                            &              &              &               &         \\
\multicolumn{3}{l}{CAPIO (ensemble) \cite{capio}} &  2.68 & 7.56 & 3.19 & 7.64             \\
\multicolumn{3}{l}{\small \emph{(Ensemble of 8 systems)}}                            &              &              &               &         \\
\multicolumn{3}{l}{DeepSpeech 2 \cite{deepspeech2}}     & - & - & 5.83 & 12.69              \\
\multicolumn{3}{l}{Sequence-to-sequence \cite{improvedzeyer}}  &3.54 & 11.52 & 3.82 & 12.76            \\ \midrule
\multicolumn{2}{c}{Front-end}                     & LM                          &              &              &               &              \\ \cmidrule{1-3} 
\multicolumn{2}{c}{Mel}                           & 4-gram  & 4.26 & 13.80 & 4.82 & 14.54\\
\multicolumn{2}{c}{Mel}                           & ConvLM & 3.13 & 10.61 & 3.45 & 11.92           \\
\multicolumn{2}{c}{Learnable (40)}                     & ConvLM & 3.16 &10.05 &3.44 &11.24            \\
\multicolumn{2}{c}{Learnable (80)}                     & ConvLM & 3.08 &9.94 &3.26 &10.47  \\     \bottomrule
\end{tabular}
\caption{WER (\%) on Librispeech.}
\label{tab:libri}
\vspace{-4mm}
\end{table}

\subsubsection{Librispeech dataset}
Table \ref{tab:libri} reports WER on the Librispeech dataset. The CAPIO \cite{capio} ensemble model combines the lattices from 8 individual HMM-DNN systems (using both convolutional and LSTM layers), and is the current state-of-the-art on Librispeech. CAPIO (single) is the best individual system, selected either on dev-clean or dev-other. The sequence-to-sequence baseline is an encoder-decoder with attention and a BPE-level \cite{bpe} LM, and currently the best end-to-end system on this dataset. We can observe that our fully convolutional model improves over CAPIO (Single) on the clean part, and is the current best end-to-end system on test-other with an improvement of $2.3\%$ absolute. Our system also outperforms DeepSpeech 2 on both test sets by a significant margin. An interesting observation is the impact of each convolutional block. While replacing the 4-gram LM by a convolutional LM improves similarly on the clean and noisier parts, learning the speech front-end gives similar performance on the clean part but significantly improves the performance on noisier, harder utterances, a finding that is consistent with previous literature \cite{hoshen2015speech}. Moreover, doubling the number of learnable filters improves the performance of our system on all development and test sets, which is consistent with our results on WSJ. 

\subsection{Analysis of the convolutional language model}
Since this paper uses convolutional language models for speech recognition systems for the first time, we present additional studies of the language model in isolation. These experiments use our best language model on Librispeech, and evaluations in WER are carried out using the baseline system trained on mel-filterbanks.  The decoder parameters are tuned using the grid search described in Section~\ref{sec:experiments}. To save computational resources, in these experiments, we use a slightly suboptimal beam size/beam score fixed to 2500/30 respectively.

\begin{figure}[t]
  \includegraphics[width=\columnwidth,height=6cm]{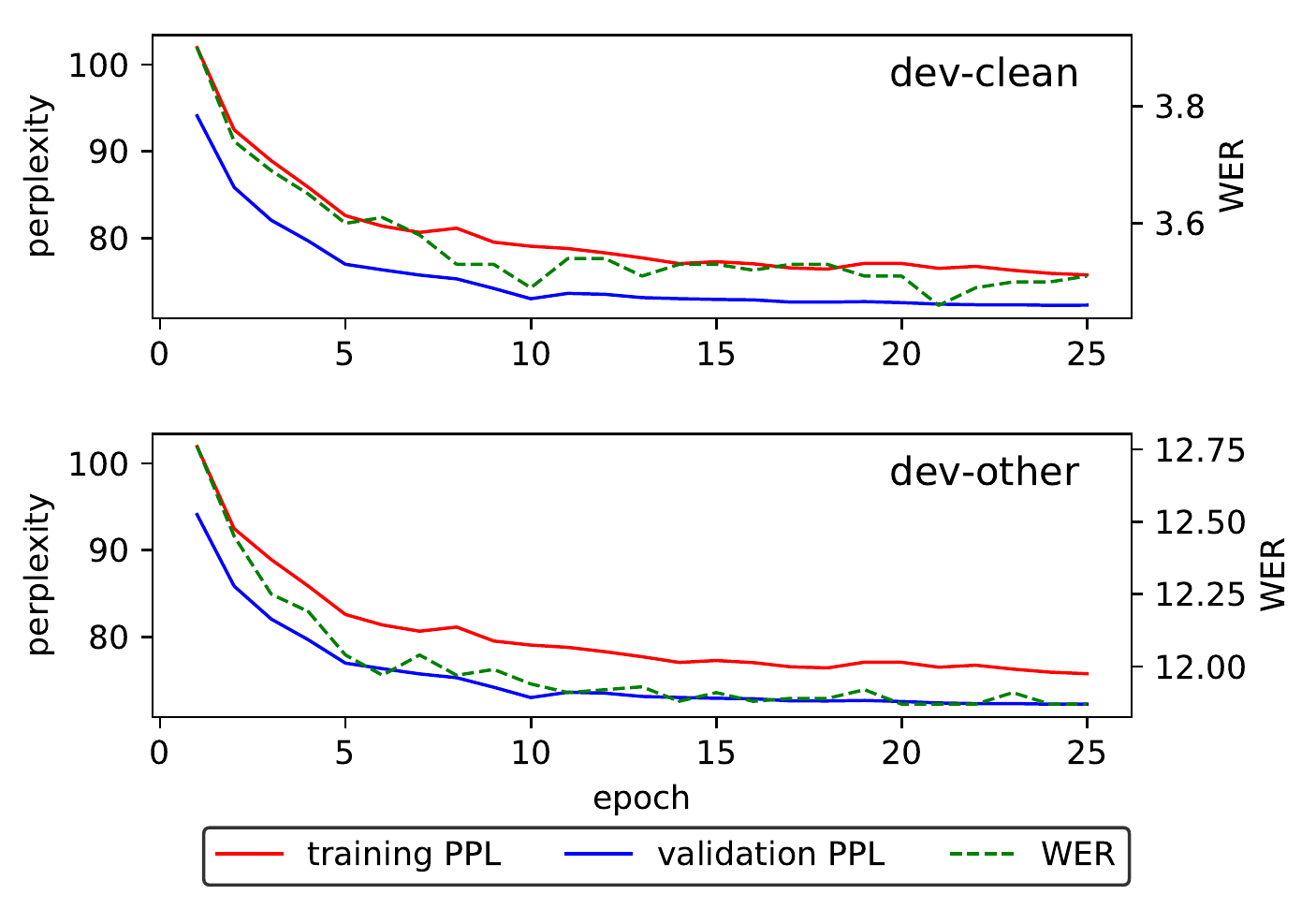}
  \caption{Evolution of WER (\%) on Librispeech with the perplexity of the language model.}
  \label{fig:ppl_wer}
\end{figure}

{\bf Perplexity and WER}
Figure \ref{fig:ppl_wer} shows the correlation between perplexity and WER as the training progresses. 
As perplexity decreases, the WER on both \emph{dev-clean} and 
\emph{dev-other} also decreases following the same trend. 
It illustrates that perplexity on the linguistic data is a good surrogate of the final performance of the speech recognition pipeline. Architectural choices or hyper-parameter tuning can thus be carried out mostly using perplexity alone.

{\bf Context and WER}
Table \ref{tab:context_wer} reports WER obtained for different context sizes used in the LM. Looking at contexts ranging from 3 (comparable to the n-gram baseline) to 50 for our best language model, the WER decreases monotonically until a context size of about $20$, and then almost stays still. We observe that the convolutional LM already improves on the n-gram model even with the same context size. Increasing the context gives a significant boost in performance, with the major gains obtained between a context of $3$ to $9$ ($-1.9\%$ absolute WER).

\begin{table}[t]
    \centering
    \begin{tabular}{cccc}
    \toprule
    \multirow{2}{*}{Model} & \multirow{2}{*}{Context} & \multicolumn{2}{c}{WER} \\
    \cmidrule{3-4}
                       &                          & dev-clean  & dev-other  \\
    \midrule 
    4-gram      & 3              & 4.26 & 13.80  \\
    \midrule
    ConvLM     & 3             & 4.11 & 13.17 \\
    ConvLM     & 9             & 3.34 & 11.29 \\
    ConvLM     & 19             & 3.27 & 11.06 \\
    ConvLM     & 29             & 3.25 & 11.09 \\
    ConvLM     & 39             & 3.24 & 11.07 \\
    ConvLM     & 49             & 3.24 & 11.08 \\
    \bottomrule
    \end{tabular}
  \caption{Evolution of WER (\%) on Librispeech with the context size of the language model.}
  \label{tab:context_wer}
  \vspace{-5mm}
\end{table}

\subsection{Analysis of the learnt front-ends}

In order to understand qualitatively the advantages of a learnable front-end over mel-filterbanks, we compare the frequency scale learned by our systems trained on the waveform. To do so, we compute a power spectrum of each convolutional filter of the first layer. We then define as its center frequency, the frequency at which its power spectrum is maximal. 

\begin{figure}[t]
  \includegraphics[width=0.9\columnwidth]{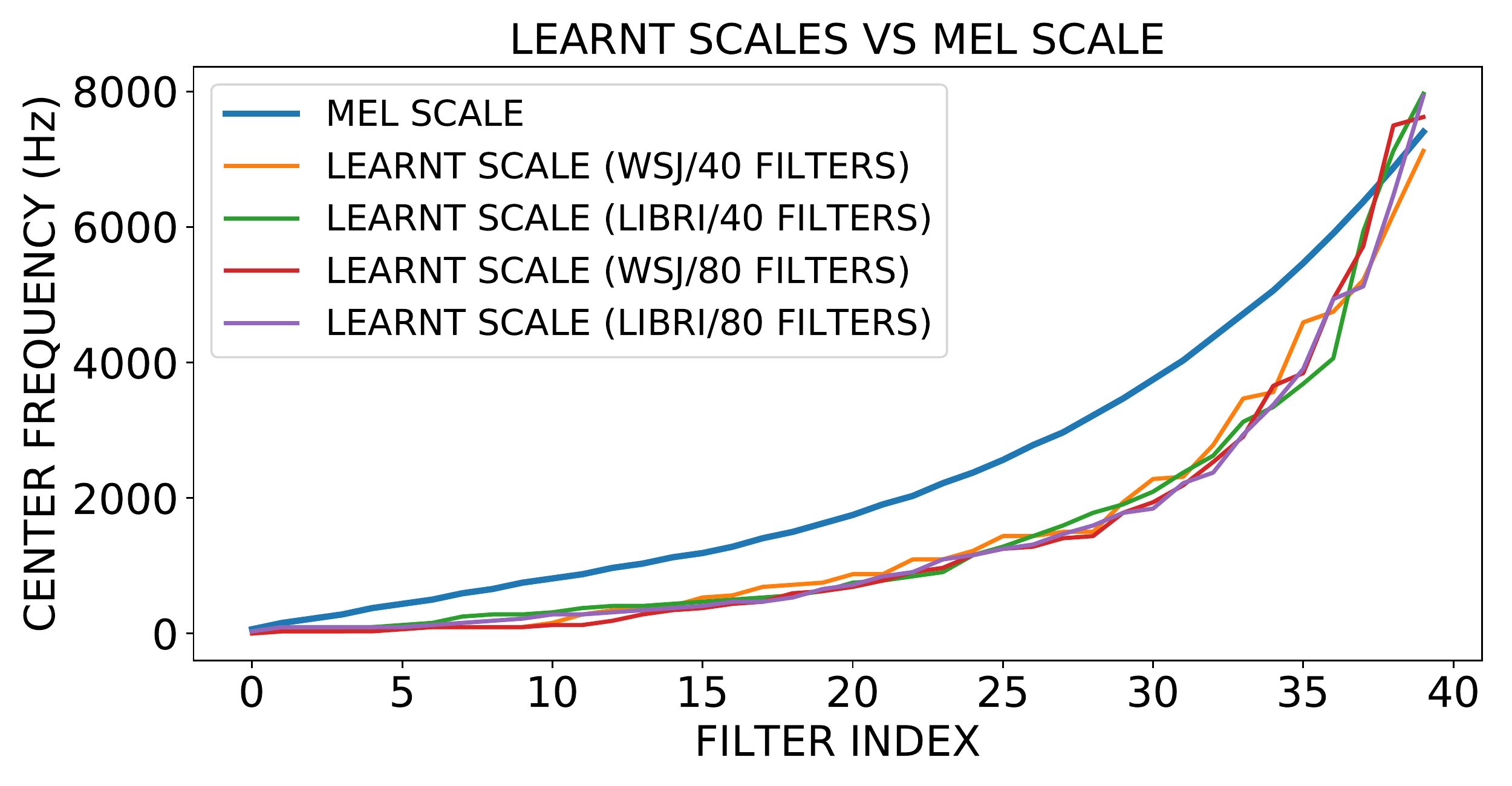}
  \caption{Center frequency of the front-end filters, for the mel-filterbank baseline and the learnable front-ends.}
  \label{fig:scale_comp}
\end{figure}

Figure \ref{fig:scale_comp} compares the mel-scale to the scales learnt by the 
front-ends of our best models on WSJ and Librispeech.
The figure is obtained by sorting the filters according to their center frequency, downsampling by a factor of $2$ for the front-ends with $80$ filters.
We observe that all 
models converge to 
similar scales, regardless of the dataset they are trained on, or how many filters they have. These learnt scales show a logarithmic shape similar to a mel-scale, but 
they are significantly more biased towards the lower frequencies. Moreover, these learnt scales closely relate to those learned by previously proposed gammatone-based learnable front-ends (see Figure 2 of \cite{hoshen2015speech} and Figure 3 of \cite{sainath2015learning}).

\begin{figure}[t]
  \includegraphics[width=\columnwidth,trim={0, 7.6cm, 0, 7.6cm},clip]{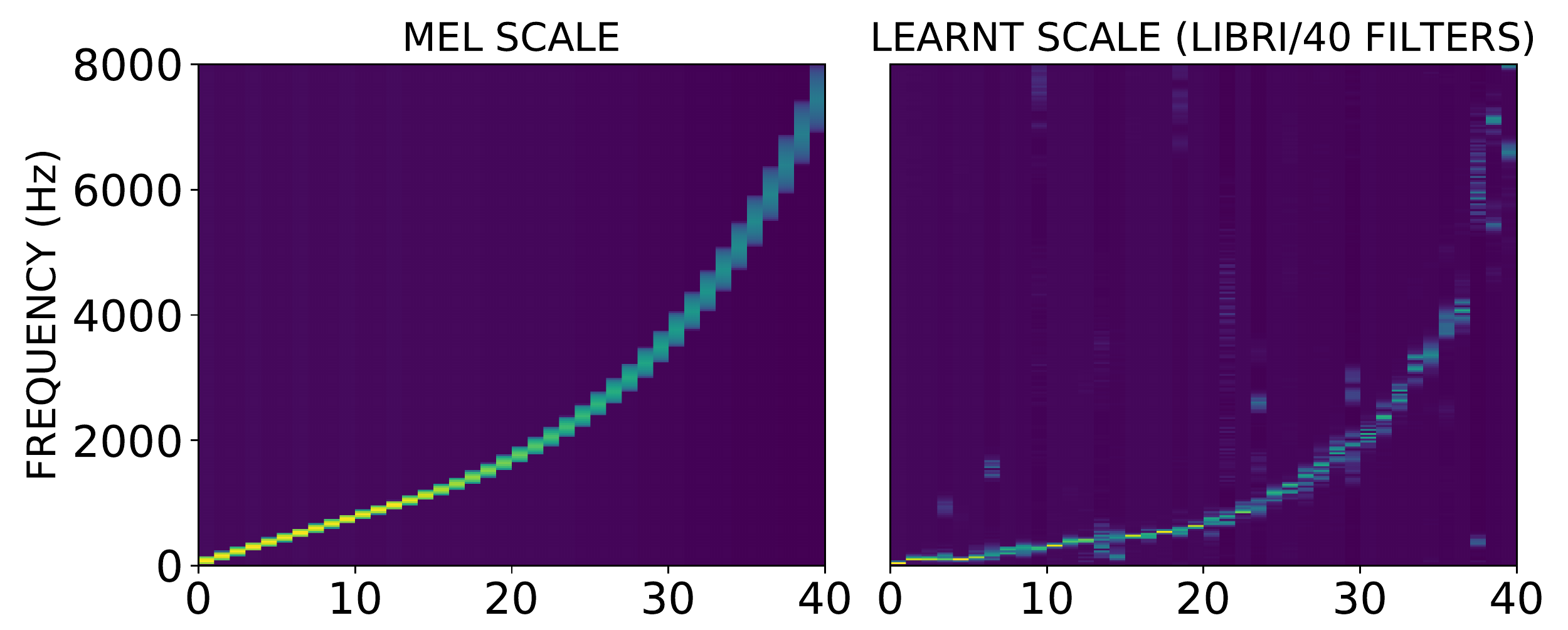}
  \caption{Power heatmap of the 40 mel-filters (left) and of the frequency response of the 40 convolutional filters learned from the raw waveform on Librispeech (right).}
  \label{fig:heat_comp}
\end{figure}

Still, analyzing a learnable filter only by its center frequency does not inform on other important characteristics such as its localization in frequency. Figure \ref{fig:heat_comp} shows a heatmap of the power of the mel filters along the frequency axis, and of the 40 filters learned on Librispeech, at convergence. We observe that even though the learnt filters show some artifacts in high frequencies, they are mostly localized around their center frequency, similarly to mel-filters. 

Overall, the consistency of the scales learned by various front-ends, over several studies, and on many datasets, suggests strongly that the mel-scale is suboptimal for automatic speech recognition and that learnable front-ends are able to learn a more appropriate one.

\section{Conclusion}

We introduced the first fully convolutional pipeline for speech recognition, that can directly process the raw waveform and shows state-of-the art performance on Wall Street Journal and on Librispeech among end-to-end systems. This first attempt at exploiting convolutional language models in speech recognition improves significantly over a $4$-gram language model on both datasets. Replacing mel-filterbanks by a learnable front-end gives additional gains in performance, that appear to be more prevalent on noisy data. This suggests learning the front-end is a promising avenue for speech recognition with challenging recording conditions.


\bibliographystyle{IEEEtran}
\bibliography{main}

\begin{thebibliography}{10}
\providecommand{\url}[1]{#1}
\csname url@samestyle\endcsname
\providecommand{\newblock}{\relax}
\providecommand{\bibinfo}[2]{#2}
\providecommand{\BIBentrySTDinterwordspacing}{\spaceskip=0pt\relax}
\providecommand{\BIBentryALTinterwordstretchfactor}{4}
\providecommand{\BIBentryALTinterwordspacing}{\spaceskip=\fontdimen2\font plus
\BIBentryALTinterwordstretchfactor\fontdimen3\font minus
  \fontdimen4\font\relax}
\providecommand{\BIBforeignlanguage}[2]{{%
\expandafter\ifx\csname l@#1\endcsname\relax
\typeout{** WARNING: IEEEtran.bst: No hyphenation pattern has been}%
\typeout{** loaded for the language `#1'. Using the pattern for}%
\typeout{** the default language instead.}%
\else
\language=\csname l@#1\endcsname
\fi
#2}}
\providecommand{\BIBdecl}{\relax}
\BIBdecl

\bibitem{glu}
Y.~Dauphin, A.~Fan, M.~Auli, and D.~Grangier, ``Language modeling with gated
  convolutional networks,'' in \emph{ICML}, 2017.

\bibitem{fairseq}
J.~Gehring, M.~Auli, D.~Grangier, D.~Yarats, and Y.~Dauphin, ``Convolutional
  sequence to sequence learning,'' in \emph{ICML}, 2017.

\bibitem{fairseq_translation}
J.~Gehring, M.~Auli, D.~Grangier, and Y.~Dauphin, ``A convolutional encoder
  model for neural machine translation,'' in \emph{ACL}, 2017.

\bibitem{wavenet}
A.~van~den Oord, S.~Dieleman, H.~Zen, K.~Simonyan, O.~Vinyals, A.~Graves,
  N.~Kalchbrenner, A.~W. Senior, and K.~Kavukcuoglu, ``Wavenet: A generative
  model for raw audio,'' in \emph{SSW}, 2016.

\bibitem{CTC}
A.~Graves and N.~Jaitly, ``Towards end-to-end speech recognition with recurrent
  neural networks,'' in \emph{ICML}, 2014.

\bibitem{RNNLM}
T.~Mikolov, M.~Karafi{\'a}t, L.~Burget, J.~Cernock{\'y}, and S.~Khudanpur,
  ``Recurrent neural network based language model,'' in \emph{INTERSPEECH},
  2010.

\bibitem{deepspeech2}
D.~Amodei, S.~Ananthanarayanan, R.~Anubhai, J.~Bai, E.~Battenberg, C.~Case,
  J.~Casper, B.~Catanzaro, Q.~Cheng, G.~Chen \emph{et~al.}, ``Deep speech 2:
  End-to-end speech recognition in english and mandarin,'' in
  \emph{International Conference on Machine Learning}, 2016, pp. 173--182.

\bibitem{seq2seqspeech}
W.~Chan, N.~Jaitly, Q.~V. Le, and O.~Vinyals, ``Listen, attend and spell,''
  \emph{CoRR}, vol. abs/1508.01211, 2015.

\bibitem{improvedzeyer}
A.~Zeyer, K.~Irie, R.~Schl{\"u}ter, and H.~Ney, ``Improved training of
  end-to-end attention models for speech recognition,'' \emph{arXiv preprint
  arXiv:1805.03294}, 2018.

\bibitem{tdnn}
A.~H. Waibel, T.~Hanazawa, G.~E. Hinton, K.~Shikano, and K.~J. Lang, ``Phoneme
  recognition using time-delay neural networks,'' \emph{IEEE Trans. Acoustics,
  Speech, and Signal Processing}, vol.~37, pp. 328--339, 1989.

\bibitem{cnn_abdelhamid}
O.~Abdel-Hamid, A.~rahman Mohamed, H.~Jiang, L.~Deng, G.~Penn, and D.~Yu,
  ``Convolutional neural networks for speech recognition,'' \emph{IEEE/ACM
  Transactions on Audio, Speech, and Language Processing}, vol.~22, pp.
  1533--1545, 2014.

\bibitem{wav2letter2}
\BIBentryALTinterwordspacing
V.~Liptchinsky, G.~Synnaeve, and R.~Collobert, ``Letter-based speech
  recognition with gated convnets,'' \emph{CoRR}, vol. abs/1712.09444, 2017.
  [Online]. Available: \url{http://arxiv.org/abs/1712.09444}
\BIBentrySTDinterwordspacing

\bibitem{zhang_towards}
Y.~Zhang, M.~Pezeshki, P.~Brakel, S.~Zhang, C.~Laurent, Y.~Bengio, and A.~C.
  Courville, ``Towards end-to-end speech recognition with deep convolutional
  neural networks,'' in \emph{INTERSPEECH}, 2016.

\bibitem{tdfbanks1}
N.~Zeghidour, N.~Usunier, I.~Kokkinos, T.~Schatz, G.~Synnaeve, and E.~Dupoux,
  ``Learning filterbanks from raw speech for phone recognition,'' \emph{2018
  IEEE International Conference on Acoustics, Speech and Signal Processing
  (ICASSP)}, pp. 5509--5513, 2018.

\bibitem{palaz2015convolutional}
D.~Palaz, M.~M. Doss, and R.~Collobert, ``Convolutional neural networks-based
  continuous speech recognition using raw speech signal,'' in \emph{Acoustics,
  Speech and Signal Processing (ICASSP), 2015 IEEE International Conference
  on}.\hskip 1em plus 0.5em minus 0.4em\relax IEEE, 2015, pp. 4295--4299.

\bibitem{hoshen2015speech}
Y.~Hoshen, R.~J. Weiss, and K.~W. Wilson, ``Speech acoustic modeling from raw
  multichannel waveforms,'' in \emph{Proceedings of ICASSP}.\hskip 1em plus
  0.5em minus 0.4em\relax IEEE, 2015.

\bibitem{sainath2015learning}
T.~N. Sainath, R.~J. Weiss, A.~Senior, K.~W. Wilson, and O.~Vinyals, ``Learning
  the speech front-end with raw waveform cldnns,'' in \emph{Interspeech}, 2015.

\bibitem{tdfbanks2}
N.~Zeghidour, N.~Usunier, G.~Synnaeve, R.~Collobert, and E.~Dupoux,
  ``End-to-end speech recognition from the raw waveform,'' in
  \emph{Interspeech}, 2018.

\bibitem{chan2015deep}
W.~Chan and I.~Lane, ``Deep recurrent neural networks for acoustic modelling,''
  \emph{arXiv preprint arXiv:1504.01482}, 2015.

\bibitem{ulyanov2017instance}
D.~Ulyanov, A.~Vedaldi, and V.~Lempitsky, ``Instance normalization: the missing
  ingredient for fast stylization,'' \emph{arXiv preprint arXiv:1607.08022},
  2017.

\bibitem{dropout}
N.~Srivastava, G.~E. Hinton, A.~Krizhevsky, I.~Sutskever, and R.~Salakhutdinov,
  ``Dropout: a simple way to prevent neural networks from overfitting.''
  \emph{Journal of machine learning research}, vol.~15, no.~1, pp. 1929--1958,
  2014.

\bibitem{wav2letter}
R.~Collobert, C.~Puhrsch, and G.~Synnaeve, ``Wav2letter: an end-to-end
  convnet-based speech recognition system,'' \emph{arXiv preprint
  arXiv:1609.03193}, 2016.

\bibitem{resnet}
K.~He, X.~Zhang, S.~Ren, and J.~Sun, ``Deep residual learning for image
  recognition,'' in \emph{Proceedings of the IEEE conference on computer vision
  and pattern recognition}, 2016, pp. 770--778.

\bibitem{wsj}
D.~B. Paul and J.~M. Baker, ``The design for the wall street journal-based csr
  corpus,'' in \emph{Proceedings of the workshop on Speech and Natural
  Language}.\hskip 1em plus 0.5em minus 0.4em\relax Association for
  Computational Linguistics, 1992, pp. 357--362.

\bibitem{librispeech}
V.~Panayotov, G.~Chen, D.~Povey, and S.~Khudanpur, ``Librispeech: an asr corpus
  based on public domain audio books,'' in \emph{Acoustics, Speech and Signal
  Processing (ICASSP), 2015 IEEE International Conference on}.\hskip 1em plus
  0.5em minus 0.4em\relax IEEE, 2015, pp. 5206--5210.

\bibitem{nag}
I.~Sutskever, J.~Martens, G.~Dahl, and G.~Hinton, ``On the importance of
  initialization and momentum in deep learning,'' in \emph{International
  conference on machine learning}, 2013, pp. 1139--1147.

\bibitem{latticefreemmi}
H.~Hadian, H.~Sameti, D.~Povey, and S.~Khudanpur, ``End-to-end speech
  recognition using lattice-free mmi,'' in \emph{Interspeech}, 2018.

\bibitem{capio}
K.~J. Han, A.~Chandrashekaran, J.~Kim, and I.~Lane, ``The capio 2017
  conversational speech recognition system,'' 2017.

\bibitem{bpe}
R.~Sennrich, B.~Haddow, and A.~Birch, ``Neural machine translation of rare
  words with subword units,'' \emph{arXiv preprint arXiv:1508.07909}, 2015.

\end{thebibliography}
\end{document}